\def\year{2019}\relax
\documentclass[letterpaper]{article} 
\usepackage{aaai19}  
\usepackage{times}  
\usepackage{helvet} 
\usepackage{courier}  
\usepackage[hyphens]{url}  
\usepackage{graphicx} 
\usepackage{graphicx}  
\usepackage[square,sort]{natbib}
\usepackage{latexsym}
\usepackage{subcaption}
\title{Design and Challenges of Cloze-Style\\Reading Comprehension Tasks on Multiparty Dialogue}
\author{Changmao Li \\
  Emory University \\
  400 Dowman Dr. \\
  Atlanta, GA 30322, USA \\
  \texttt{changmao.li@emory.edu} \\\And
  Tianhao Liu \\
  Emory University \\
  400 Dowman Dr. \\
  Atlanta, GA 30322, USA \\
  \texttt{tliu236@emory.edu} \\\And
  Jinho D. Choi \\
  Emory University \\
  400 Dowman Dr. \\
  Atlanta, GA 30322, USA \\
  \texttt{jinho.choi@emory.edu} \\}
 
\begin{document}

\maketitle

\begin{abstract}
This paper analyzes challenges in cloze-style reading comprehension on multiparty dialogue and suggests two new tasks for more comprehensive predictions of personal entities in daily conversations.
We first demonstrate that there are substantial limitations to the evaluation methods of previous work, namely that randomized assignment of samples to training and test data substantially decreases the complexity of cloze-style reading comprehension. According to our analysis, replacing the random data split with a chronological data split reduces test accuracy on previous single-variable passage completion task from 72\% to 34\%, that leaves much more room to improve.
Our proposed tasks extend the previous single-variable passage completion task by replacing more character mentions with variables. 
Several deep learning models are developed to validate these three tasks. A thorough error analysis is provided to understand the challenges and guide the future direction of this research.
\end{abstract}

\section{Introduction}
Reading comprehension is an intriguing task that assesses a machine's ability in understanding evidence contexts through question answering. Most previous work in reading comprehension has focused on either formal documents \cite{Rajpurkar:2016a} or children's stories \cite{Richardson:2013a}. Only few approaches have attempted comprehension on multiparty dialogue \cite{Ma:2018a}. However, with the explosive expansion of social media, data on dialogue has become dominant on the web. Inspired by various options of analytic models and the potential of the dialogue processing market, we extend the corpus presented by \citet{Ma:2018a} for comprehensive predictions of personal entities in multiparty dialogue and develop deep learning models to make robust inference on their contexts.

\noindent Passage completion on multiparty dialogue is one of the reading comprehension tasks that requires a model to match the conversational dialogues with the formal (passage) writings. By building a robust model for this task, people can tell the status of their favorite characters by checking synopsis to see if their favorite characters appear in specific episodes without watching the entire series, thus allowing them an efficient way to decide whether to watch a particular episode. Involving matching contexts between colloquial (dialog) and formal (passage) writings makes this task extremely challenging. 

Distinguished from the previous work that only focused on a single variable per passage \cite{Ma:2018a},  we propose two new passage completion tasks on multiparty dialogue which increase the task complexity by replacing more character mentions with variables with a better motivated data split. The details of these tasks are in Section \ref{tasksrefs}. Several deep neural network models are developed to validate the three reading comprehension tasks. Based on the experimental results, we aim to identify main challenges in these tasks and suggest future research directions. 

\section{Related Work}

The CNN/Daily Mail dataset introduced by \citet{Hermann:2015a} is to infer the missing entity (answer \textit{a}) of a question \textit{q} from all the possible entities which appear in a passage \textit{p} while the passage is a news article, the question is a cloze-style task, in which one of entities is replaced by a placeholder, and the answer is this questioned entity. Many other models have been proposed to tackle this dataset, which are either RNN and attention based \cite{Chen:2016a} or CNN and RNN based \cite{Trischler:2016a} or gated attention based \cite{Dhingra:2017a} or attention over attention based \cite{Cui:2017a}. Finally, the pretrained bi-directional transformer encoder(BERT) which was introduced by \citet{Devlin:2018a} shows that the pretraining of the language representations can bring better performance in several downstream tasks including machine reading comprehension. More datasets are available for another type of a reading comprehension task, that is multiple choice question answering, such as MCTest \cite{Richardson:2013a} , TriviaQA \cite{Joshi:2017a}, RACE \cite{Lai:2017a} and Dream \cite{Sun:2019a}.

Unlike the above tasks where documents and queries are written in a similar writing style, the multiparty dialogue reading comprehension task introduced by \citet{Ma:2018a} has a very different writing style between dialogues and queries. However, their randomized assignment of samples to  training and test data substantially decreases the complexity of cloze-style reading comprehension. In this paper, we address this issue by introducing a chronological data split and new variants of the cloze-style reading comprehension task to challenge even higher task complexity.

\section{Corpus}

Our corpus comes from the transcripts of the TV show \textit{Friends} with ten seasons collected by the Character Mining project.\footnote{\url{https://github.com/emorynlp/character-mining}} Each season contains about 24 episodes, each episode is split into about 13 scenes, where each scene comprises a sequence of about 21 utterances. This dataset contains several layers of annotation. The first two seasons of the show for an entity linking task was annotated by \citet{Chen:2016b}. Plot summaries of all episodes for the first eight seasons were collected by \citet{Jurczyk:2017a}  to evaluate a document retrieval task. The rest of the plot summaries were collected by \citet{Ma:2018a}. Table \ref{table:corpus} shows the statistical data of the corpus from \citet{Ma:2018a}.

\begin{table}[ht]
\begin{center}
\centering\resizebox{\columnwidth}{!}{
\begin{tabular}{l||l}
\bf Type                               & \bf Count             \\ \hline \hline
\# of dialogs                       & 1,681        \\
\# of plots                         & 4,646        \\
Avg. \# of utterances per dialog    & 15.8         \\
Avg. \# of tokens per dialog/plot   & 290.8 / 19.9 \\
Avg. \# of mentions per dialog/plot & 24.4 / 3.0   \\
Avg. \# of entities per dialog/plot & 5.4 / 2.2    \\
Max \# of entities per dialog/plot  & 16 / 7      
\end{tabular}}
\end{center}
\caption{\label{table:corpus} Corpus statistical data from \citet{Ma:2018a}}
\end{table}

\noindent Based on the above corpus we created a new data split different from \citet{Ma:2018a}'s data split. In the previous work of \citet{Ma:2018a}, they used a random data split where 1,187 of 1,349 queries in the development set and 1,207 of 1,353 queries in the test set are generated from the same plot summaries as some queries in the training set with only masking the different character entities which makes the model can see the right answer in the training set. To fix this issue, we created a new data split, the detail of which is in Section \ref{datasplitref}.

\section{Tasks} \label{tasksrefs}

Table \ref{table:dialogue} and Table \ref{table:plot} show an example of a dialogue and its plots.We propose three tasks, one is from \citet{Ma:2018a}, and another two tasks are new tasks designed by us. 

\begin{table*}[ht!] 
    \begin{subtable}{\textwidth}
    \caption{A dialogue from \textit{Friends}: Season 1, Episode 21, Scene 1.}
    \label{table:dialogue}
    \resizebox{\textwidth}{!}{\begin{tabular}{l|l|l}
    \textbf{ID} & \textbf{Speaker} & \textbf{Utterance} \\ \hline \hline
    1 & \textbf{@ent04} & How could someone get a hold of your credit card number ? \\
    2 & \textbf{@ent00} & I have no idea . But look how much they spent ! \\
    3 & \textbf{@ent01} & \textbf{@ent00} , would you calm down ? The credit card people said that you only have to pay for the stuff that you bought . \\
    4 & \textbf{@ent00} & I know . It 's just such reckless spending . \\
    5 & \textbf{@ent06} & I think when someone steals your credit card , \\
      &  & they 've kind of already thrown caution to the wind . \\
    6 & \textbf{@ent05} & Wow , what a geek . They spent \$ 69.95 on a Wonder Mop . \\
    7 & \textbf{@ent00} & That 's me . \\
    ...&...&...
    \end{tabular}}
    \end{subtable}
    \vspace{2ex}
    \begin{subtable}{\textwidth}
    \caption{Plots generated for the dialog in(a)}
    \label{table:plot}
    \begin{center}
    \resizebox{\textwidth}{!}{
    \begin{tabular}{l|l}
    \textbf{ID} & \textbf{Plots} \\ \hline \hline
    1 & \textbf{@ent00} spent \$ 69.95 on a Wonder Mop \\
    2 & \textbf{@ent04} asks \textbf{@ent00} how someone could get a hold of \textbf{@ent00} 's credit card number and \textbf{@ent00} is surprised at how much was spent . \\
    \end{tabular}}
    \end{center}
    \end{subtable}
    \vspace{2ex}
    \begin{subtable}{\textwidth}
    \caption{Queries generated from the passages in (b) for single variable task}
    \label{figure:task1}
    \resizebox{\textwidth}{!}{
    \begin{tabular}{l|l}
    \textbf{ID} & \textbf{Queries} \\ \hline \hline
    1.a & \textbf{\textit{x}} spent \$ 69.95 on a Wonder Mop \\
    2.a & \textbf{\textit{x}} asks \textbf{@ent00} how someone could get a hold of \textbf{@ent00} 's credit card number and \textbf{@ent00} is surprised at how much was spent . \\
    2.b & \textbf{@ent04} asks \textbf{\textit{x}} how someone could get a hold of \textbf{@ent00} 's credit card number and \textbf{@ent00} is surprised at how much was spent . \\
    2.c & \textbf{@ent04} asks \textbf{@ent00} how someone could get a hold of \textbf{\textit{x}} 's credit card number and \textbf{@ent00} is surprised at how much was spent . \\
    2.d & \textbf{@ent04} asks \textbf{@ent00} how someone could get a hold of \textbf{@ent00} 's credit card number and \textbf{\textit{x}} is surprised at how much was spent . \\
    \end{tabular}}
    \end{subtable}
    \vspace{2ex}
    \begin{subtable}{\textwidth}
    \caption{Queries generated from the passages in (b) for multiple variables task on the same entity}
    \label{figure:task2}
    \resizebox{\textwidth}{!}{
    \begin{tabular}{l|l}
    \textbf{ID} & \textbf{Queries} \\ \hline \hline
    1.a & \textbf{\textit{x}} spent \$ 69.95 on a Wonder Mop \\
    2.a & \textbf{\textit{x}} asks \textbf{@ent00} how someone could get a hold of \textbf{@ent00} 's credit card number and \textbf{@ent00} is surprised at how much was spent . \\
    2.b & \textbf{@ent04} asks \textbf{\textit{x}} how someone could get a hold of \textbf{\textit{x}} 's credit card number and \textbf{\textit{x}} is surprised at how much was spent . \\
    \end{tabular}}
    \end{subtable}
    \begin{subtable}{\textwidth}
    \caption{Queries generated from the passages in (b) for two variables task }
    \label{figure:task3}
    \resizebox{\textwidth}{!}{
    \begin{tabular}{l|l}
    \textbf{ID} & \textbf{Queries} \\ \hline \hline
    1.a & \textbf{\textit{x1}} spent \$ 69.95 on a Wonder Mop \\
    1.b & \textbf{\textit{x2}} spent \$ 69.95 on a Wonder Mop \\
    2.a & \textbf{\textit{x1}} asks \textbf{\textit{x2}} how someone could get a hold of \textbf{@ent00} 's credit card number and \textbf{@ent00} is surprised at how much was spent . \\
    2.b & \textbf{\textit{x2}} asks \textbf{\textit{x1}} how someone could get a hold of \textbf{@ent00} 's credit card number and \textbf{@ent00} is surprised at how much was spent . \\
    2.c & \textbf{\textit{x1}} asks \textbf{@ent00} how someone could get a hold of \textbf{\textit{x2}} 's credit card number and \textbf{@ent00} is surprised at how much was spent . \\
    2.d & \textbf{\textit{x2}} asks \textbf{@ent00} how someone could get a hold of \textbf{\textit{x1}} 's credit card number and \textbf{@ent00} is surprised at how much was spent . \\
    2.e & \textbf{@ent04} asks \textbf{\textit{x1}} how someone could get a hold of \textbf{\textit{x2}} 's credit card number and \textbf{@ent00} is surprised at how much was spent . \\
    ... & ...
    \end{tabular}}
    \end{subtable}
    \caption{An dialogue and its plots from the corpus and generated queries for each task. \textbf{\textit{x}}, \textbf{\textit{x1}} and \textbf{\textit{x2}} denote unknown variables.}
    \label{table:task example}
\end{table*}

\subsection{Single Variable task(SV)}
The single variable task from \citet{Ma:2018a} consists a dialogue passage \textit{p}, a query \textit{q} which is from plot summary of the dialogue passage and an answer \textit{a}. In this task, a query \textit{q} replaces only one character entity with an unknown variable \textit{x} and the machine is asked to infer the replaced character entity (answer \textit{a}) from all the possible entities appear in the dialogue passage \textit{p}. This task is evaluated by computing the accuracy of predictions (see Section \ref{evalmetric}). Table \ref{figure:task1} shows an example of this task. 

\subsection{Multiple Variable task on the Same entity(MVS)}

Similar to the first task, the multiple variable task on the same entity also consists a dialogue passage \textit{p}, a query \textit{q} which is from plot summary of the dialogue passage and an answer \textit{a}. The difference is that the query \textit{q} replaces all the same character entities with the same unknown variable \textit{x} and the machine is asked to infer this same character entity (answer \textit{a}) from all the possible entities appear in the dialogue passage \textit{p}. The difficulty of this task is that the machine has to predict a character entity who does not appear in the query. The evaluation of this task is the same as the first task since the model only predicts one entity (see Section \ref{evalmetric}). Table \ref{figure:task2} shows an example of this task. 

\subsection{Two Variables task(TV)}

To make this task more intriguing, we propose two new tasks. This task consists a dialogue passage \textit{p}, a query \textit{q} which is from plot summary of the dialogue passage and an answer pair $a_1$ and $a_2$. Here the query replaces two character entities with two different unknown variables $x_1$ and $x_2$ and the machine is asked to infer these two missing character entities(the answer pair $a_1$ and $a_2$) from all the possible entities appear in the dialogue passage \textit{p}. Two replaced character entities can be the same or different and to avoid bias between two unknown variables, the different order of the variables will be considered as different queries. Since there are several plot summaries which only have one character entity in it, the model needs to be designed to predict one or two entities, evaluated by the F1 score (Section \ref{evalmetric}).  

\section{Approaches}

We selected some published approaches which is to solve dialogue reading comprehension to validate on our new data split and three tasks.

\subsection{BiLSTM}

\begin{figure}[ht]
\centering
\includegraphics[width=\columnwidth]{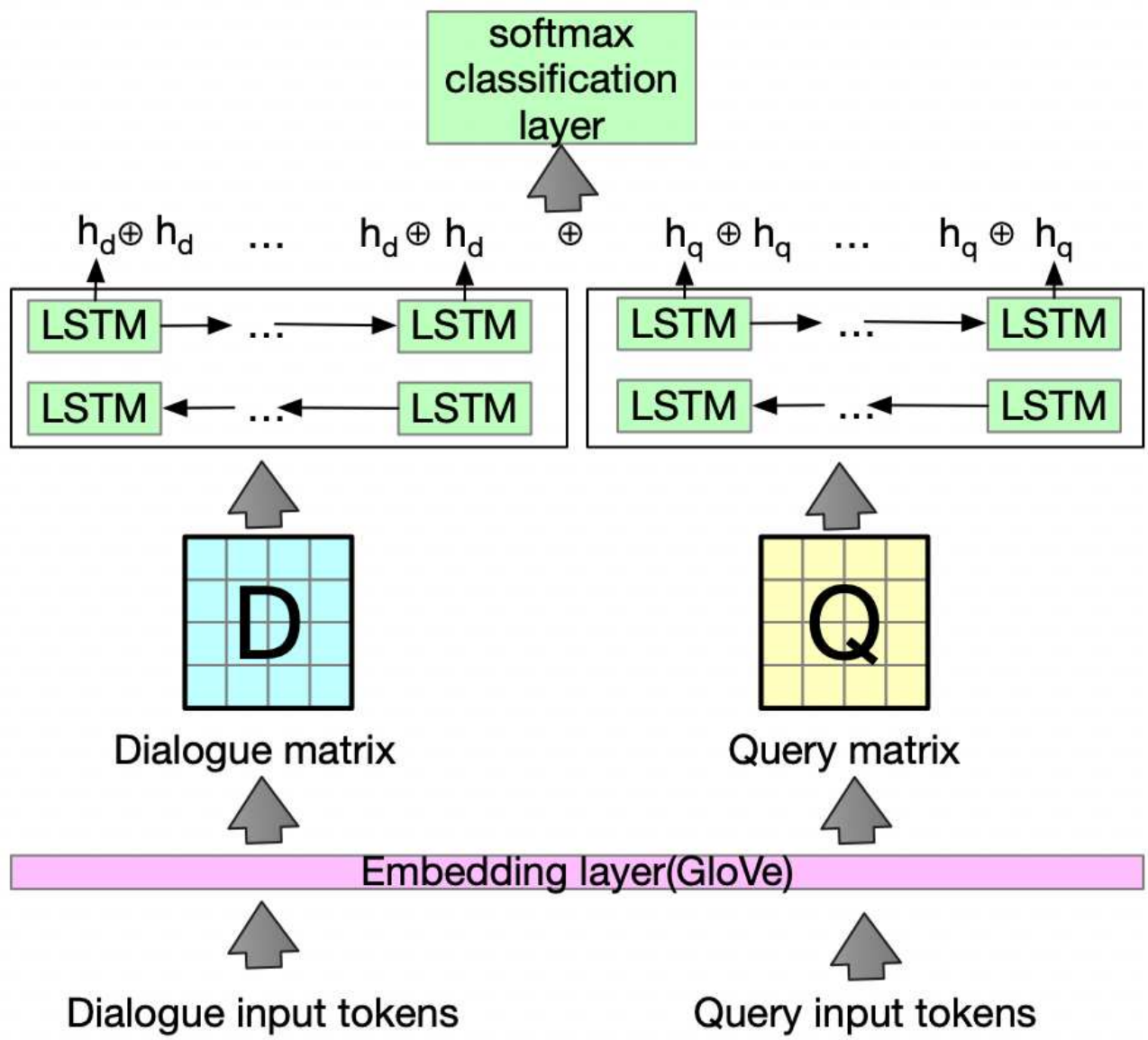}
\caption{Architecture for BiLSTM model}
\label{figure:bilstm}
\end{figure}

The LSTM model which is selected as one of the baselines methods in this project. We deal with utterances and query separately. First we generate the utterances and query embedding. For utterances we combine them into document-level matrix and input them into Bidirectional LSTM and get the output $h_d$. For query, we directly input them into BiLSTM and get the output $h_q$. Finally we concatenate the output of $h_d$ and $h_q$ and pass them into a softmax classification layer. By using Bidirectional LSTM, we can extract the sequence feature of utterances and queries. Figure \ref{figure:bilstm} shows the architecture for this model.

\subsection{CNN+BiLSTM}

Based on \citet{Ma:2018a}, we first use CNN to extract the gram-level features of utterances and then use LSTM to capture the sequence feature of both utterances and query. CNN is applied into each utterance to extract more token level n-gram features then we take max pooing and combine them into a document-level matrix, and input it into the Bidirectional LSTM and get the output $h_q$. For query, we directly into them into Bidirectional LSTM and get the output $h_q$, too. Finally we concatenate the output of $h_d$ and $h_q$ and pass them into a softmax classification layer. By using cnn and Bidirectional LSTM theoretically we can both capture the n-grams features and sequence features.

\subsection{CNN+BiLSTM+UA+DA}

This method is the SOTA method last year in \citet{Ma:2018a}'s data split which is also selected as one of our experimental methods. Similar to the last one, they still use CNN to extract token-level n-gram features of utterances and use LSTM to capture the sequence feature, however, they added the utterance-level attention and document-level attention to extract more features related to similarity between query and utterances. This utterance-level attention basically is to compute the similarity between each utterance and query. This document level attention is to compute the similarity between query and whole document matrix. Their output is concatenated with original Bidirectional LSTM output and pass them into a softmax classification layer. By adding the attention layers, theoretically we can capture the some similarity features between utterances and query.

\subsection{BERT+Fine-tuning}

\begin{figure}[ht]
\centering
\includegraphics[width=\columnwidth]{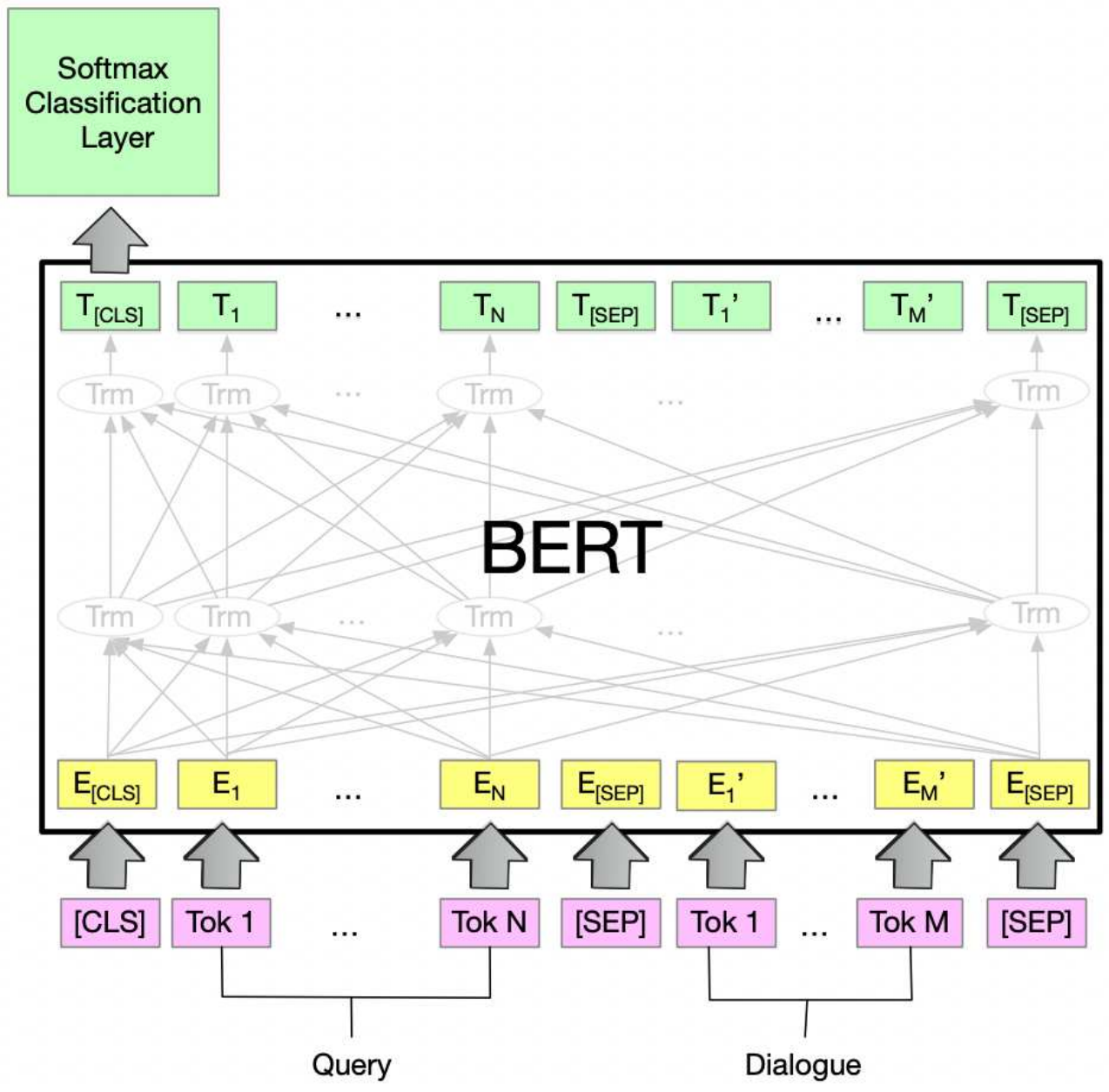}
\caption{Architecture for BERT+Fine-tuning model}
\label{figure:bert}
\end{figure}

Since BERT model \cite{Devlin:2018a} shows very wide usage in the NLP field, in this paper, we also use the BERT fine-tuning model as one of our exploratory methods. Our input format for BERT is [CLS]query[SEP]utterances[SEP]. We use BERT$_{{\rm BASE}}$ for fine-tuning since we do not have large memory of GPUs to run BERT$_{{\rm LARGE}}$. What we do is to simply input the [CLS] representation of BERT final output to our top layer. Our top layer is a softmax classification layer $W \in R^{K \times H} $, where $H$ is the dimension of the [CLS] representation and $K$ is the number of entity labels. Figure \ref{figure:bert} shows the architecture for this model.

\section{Experiments}

\subsection{Data Split Overview} \label{datasplitref}

Our data split has three parts: for each season, we use episodes 1-18 as training data, episodes 19-21 as development data, and the rest of episodes as test data. This data split is chosen to mimic the training data which a model would use in an application setting for reading comprehension, where historic data is used for training while recent data is the subject for prediction. Table \ref{table:data_split} shows the dataset split for our experiments that roughly gives 80/10/10\% for training/development/test sets. 

\begin{table}[htbp!]
\centering\small
\begin{tabular}{c||r|r|r}
\bf Task  & \bf Training & \bf Development &\bf  Test \\ \hline \hline
SV   & 11,104   & 1,486        & 1,242 \\ 
MVS  &  8,211   & 1,099        &   916  \\ 
TV   & 22,332   & 3,174        & 2,612
\end{tabular}
\caption{The number of queries in each dataset.}
\label{table:data_split}
\vspace{-1ex}
\end{table}

\subsection{Evaluation metrics} \label{evalmetric}

We use different evaluation metrics for different tasks. For the SV task and the MVS task: we use simple accuracy since the model only predicts one variable:
\begin{equation}\label{(1)}
score =\frac{C_r}{C_t}
\end{equation}
Where $C_r$ is the number of the right predictions and $C_t$ is the number of the total predictions.

For the TV task: we use F1 score, since model predicts one variable or two variables:
\begin{equation}\label{(2)}
precision= \frac{C_r}{C_a}
\end{equation}
\begin{equation}\label{(3)}
recall = \frac{C_r}{C_g}
\end{equation}
\noindent
\begin{equation}\label{(4)}
F1 = 2 \times \frac{precision*recall}{precision+recall}
\end{equation}
Where $C_r$ is the number of the right predictions and $C_a$ is the number of the actual predictions and $C_g$ is the number of the gold answer.

\subsection{Results and error analysis}

\subsubsection{Results}

Table \ref{tab:result} shows the results of our experiment. 
BiLSTM is good at capturing the sequence information of sentences; however, since it only finds some kind of answer distributions on the sequence information, it cannot capture the information of the relation between query and utterance.  

\begin{table}[htp!]
  \centering\centering\resizebox{\columnwidth}{!}{
    \begin{tabular}{l||l|l}
     \textbf{Task}     & \multicolumn{2}{||c}{\textbf{Score}} \\ \hline \hline
    \textbf{SV} & \textbf{Dev} & \textbf{Test} \\ \hline \hline
    BiLSTM & 36.72($\pm$0.38) & \textbf{33.69($\pm$0.90)}  \\
    CNN+BiLSTM & 31.11($\pm$0.93) & 31.65($\pm$0.25)  \\
    CNN+BiLSTM+UA+DA & 30.32($\pm$1.72) & 29.10($\pm$0.64)  \\
    BERT+Fine-tuning & 32.30($\pm$0.66) & 29.83($\pm$0.75)   \\ \hline \hline
    \textbf{MVS} & \textbf{Dev} & \textbf{Test} \\ \hline \hline
    BiLSTM & 50.26($\pm$0.47) & \textbf{52.68($\pm$0.63)}  \\
    CNN+BiLSTM & 46.82($\pm$2.59) & 44.98($\pm$2.18) \\
    CNN+BiLSTM+UA+DA & 49.02($\pm$1.27) & 51.92($\pm$0.29)  \\ 
    BERT+Fine-tuning & 48.39($\pm$4.17) & 47.51($\pm$4.25)  \\\hline \hline
    \textbf{TV} & \textbf{Dev} & \textbf{Test}  \\ \hline \hline
    BiLSTM & 31.30($\pm$0.52) & 22.13($\pm$1.45)  \\
    CNN+BiLSTM & 27.92($\pm$0.03) & 19.97($\pm$1.43)  \\
    CNN+BiLSTM+UA+DA & 28.55($\pm$0.13) & 21.62($\pm$2.92)  \\
    BERT+Fine-tuning & 29.50($\pm$0.83) & \textbf{30.92($\pm$0.63)} \\
    \end{tabular}}
    \caption{Results from all models.}
 \label{tab:result}
  \vspace{-2ex}
\end{table}

Adding a CNN can achieve even lower accuracy because passing sequences to the CNN only keeps important information after the pooling operation, but for dialogue data, most of the time the replaced entity needs to be decided by so much original context, when using CNN, the model loses some original context.

Utterance-level attention and document-level attention proposed by \citet{Ma:2018a} are not helpful for these tasks on our data split because dialogues contain so many informal expressions and the size of the corpus is small. With limited data size, the attention mechanism does not perform well to match different expressions with the same meaning. 

The BERT model only uses attention mechanism to model the language feature, however, the pre-trained model is from formal text(wikipeida, etc) not from dialogue corpus, so it also performs poorly on our data split and tasks.  

A counter-intuitive part is that the MVS task appears to be so much easier than the SV task. The main reason is because the model can easily learn that the entity that appears in the query will not appear in the answer which will reduce many answer choices. Another counter-intuitive part is that BiLSTM without any additive method is the best model in two of the three tasks. The main reason of this is because the attention mechanism does not work well for these tasks on this dataset.

\subsubsection{Error analysis}

For further research, we extracted 100 samples from test set of each task on their best performing model to analyze what kind of error may occur. We found that the three main types of errors for these tasks are: Hidden Meaning, Utterances Reasoning \& Summary as well as Coreference Resolution. Table \ref{table:error_type} shows the error types distribution of each task. Table \ref{table:error_example} shows the examples of three main error types.

\begin{table}[htbp!]
\centering
\centering\resizebox{\columnwidth}{!}{
\begin{tabular}{l||r|r|r}
\textbf{Error types}   & \textbf{SV} & \textbf{MVS} & \textbf{TV} \\ \hline \hline
\textbf{Hidden meaning} & 22\%   &  \textbf{28\%}  & \textbf{35\%} \\ 
\textbf{Utterance reasoning and summary} & \textbf{30\%}  & 27\%  & 24\%  \\ 
\textbf{Coreference resolution} & 18\%    & 22\%   & 22\% \\ 
Object Linking & 14\%     & 11\%        & -  \\ 
Annotation & 7\%    & 6\%        & 9\% \\
Handle single variable & -    & -        & 5\% \\
Miscellaneous &9\% & 6\% &5\% 
\end{tabular}}
\caption{Percentage of error types occurred in each task} 
\label{table:error_type}
\end{table}

\textbf{Hidden Meaning} Hidden meaning is one of the most frequent error types and it happens when a query uses formal expressions to represent the same meaning as an utterance with informal expressions. This type of error also happens when a person uses an expression but with another meaning behind it. The model cannot find the right position of the corresponding entities.

\textbf{Utterances Reasoning and Summary} Utterances reasoning and summary is another frequent error type. This type of error occurs when the model needs to predict an entity based on several actions in the query which correspond to several continuous utterances in the dialogue. We noticed that the model also needs to infer the result or effect of continuous or noncontinuous actions happening in several utterances to predict correct entities.

\begin{table}[htb]
\begin{subtable}{0.5\textwidth}
\caption{Error example for hidden meaning: "you two together" = "have a relationship"}
\resizebox{\textwidth}{!}{\begin{tabular}{l|l}
\textbf{Answer} & \textbf{Query} \\ \hline \hline
\textbf{@ent01} & \textbf{x} doesn't want to have a relationship with\textbf{@ent04} again.\\ \hline \hline
\textbf{Speaker} & \textbf{Utterance} \\ \hline \hline
... & ... \\
\textbf{@ent03} & I 'm sorry , for the last time , \\
   &why aren't you two together again ? ( silence from \textbf{@ent01} . ) \\
   &no , I know . I know , because you 're not in that place . \\
   &which would be fine , except you totally are .\\
... & ... \\
\end{tabular}}
\vspace{2ex}
\end{subtable}

\begin{subtable}{0.5\textwidth}
\caption{Error example for utterances reasoning and summary: inference of "with" who?}
\resizebox{\textwidth}{!}{\begin{tabular}{l|l}
\textbf{Answer} & \textbf{Query} \\ \hline \hline
\textbf{@ent02} & \textbf{@ent03} was with x picking out the ring. \\ \hline \hline
\textbf{Speaker} & \textbf{Utterance} \\ \hline \hline
... & ... \\
\textbf{@ent03} & what ? \\
\textbf{@ent04} & \textbf{@ent02} 's gonna ask \textbf{@ent05} to marry him ! \\
\textbf{@ent03} & oh I know , I helped pick out the ring . \\
- & ( \textbf{@ent02} laughs , turns , and sees that \textbf{@ent04} and \\
&\textbf{@ent00} aren't happy . ) \\
... & ... \\
\end{tabular}}
\vspace{2ex}
\end{subtable}

\begin{subtable}{0.5\textwidth}
\caption{Error example for coreference resolution: resolution of "he"}
\end{subtable}
\resizebox{0.5\textwidth}{!}{\begin{tabular}{l|l}
\textbf{Answer} & \textbf{Query} \\ \hline \hline
\textbf{@ent05} & \textbf{@ent01} tries to leave the set , but \textbf{@ent02} tells him that \\
&so long as \textbf{x} is conscious and present on set , \\
&they will continue to shoot the film . \\ \hline \hline
\textbf{Speaker} & \textbf{Utterance} \\ \hline \hline
...&...\\
\textbf{@ent01} & \textbf{@ent03} ! you got ta let me go . the guy 's hammered ! \\
\textbf{@ent02} & I 'm sorry \textbf{@ent01} , as long as he 's here and he 's conscious \\
&we 're still shooting . \\
- & ( he walks away and \textbf{@ent01} does \textbf{@ent04} 's fist thing . \\
&he then enters \textbf{@ent05} 's dressing room , \\ 
& to find \textbf{@ent05} cutting his steak with his sword . ) \\
\textbf{@ent05} & you would n't happen to have a very big fork ?\\
...&... \\
\end{tabular}}
\caption{Error examples for three main error types}
\label{table:error_example}
\end{table}

\textbf{Coreference Resolution} This type of error occurs when there are coreferences in the utterances and query and it is confusing to determine what is the correct entity behind the pronoun. This type of error occurs when there are numerous entities in the dialogue and the model cannot differentiate them because of less recognition of coreferences.

\textbf{Object Linking} This type of error refers to a situation when a query needs to know which entity has some kind of relationship with some kind of object(for example, another entity, substance, etc.). It can be especially confusing when the object and entity are not in the same utterance.

\textbf{Annotation Issues} This type of error refers to some issues when doing annotation, such as mislabeling of answers, unanswerable or irrelevant queries, misspellings in the utterance, mislabeling the entity in the dialogues, etc. This type of error is trivial and hard to solve but still reflects the real situation in human communication.

\textbf{Handle Single Variable} This type of error occurs when solving the TV task since the TV task can either have a single variable or two variables. The model cannot tell the differences between the single variable and two variables, and most of the time this occurs when the correct answer is only one variable, the model predicts two variables.

\textbf{Miscellaneous}  This type of error represents error samples in which we cannot find apparent reason for the wrong prediction. We need to research more to find them.

\section{Conclusion and Future Work}
In this paper, we addressed the issue with the evaluation method of previous work by introducing a new data split and introduced new variants of the cloze-style reading comprehension task to increase task complexity. In addition, we ran several neural network models to validate our new data split and new tasks. By introducing evaluation methods that better motivate the dialogue reading comprehension task, we discovered many challenges that the model cannot easily tackle in the real situation. Finally, we propose what we will do in the next step to address some of these challenges.

In order to understand the hidden meaning in the dialogues, we plan to create dialogue specific meaning representation to enable the model to learn the hidden meaning of what the entity is trying to express. However, this kind of approach is also limited by its labeled data size. We will try to find unsupervised ways to enable the model to learn the semantic information itself. Additionally, we will intrigue more approaches to enable model's understanding of hidden meaning in the dialogue. For example, we will design more appropriate unsupervised pre-training tasks targeting to enable the model to learn the corresponding hidden meaning of expressions which have hidden meaning.

Furthermore, to overcome reasoning and summary challenges, we plan to incorporate knowledge base reasoning by either constructing a knowledge graph or using more dialogue data to design pre-training tasks to enable the model's ability to do inference from dialgoues. Another approach we plan to try is to use a neural network model such as graphic convolution network \cite{Defferrard:2016a} that can represent a knowledge graph structure based on dialogue dataset.

\bibliographystyle{aaai}
\bibliography{main} 
\end{document}